# Non-Volatile Memory Array Based Quantization- and Noise-Resilient LSTM Neural Networks

Wen Ma
Western Digital Research
Milpitas, CA USA
wen.ma@wdc.com

Pi-Feng Chiu
Western Digital Research
Milpitas, CA USA
pi-feng.chiu@wdc.com

Won Ho Choi
Western Digital Research
Milpitas, CA USA
won.ho.choi@wdc.com

Minghai Qin
Western Digital Research
Milpitas, CA USA
minghai.qin@wdc.com

Daniel Bedau
Western Digital Research
San Jose, CA USA
daniel.bedau@wdc.com

Martin Lueker-Boden
Western Digital Research
Milpitas, CA USA
martin.lueker-boden@wdc.com

*Abstract*—In cloud and edge computing models, it is important that compute devices at the edge be as power efficient as possible. Long short-term memory (LSTM) neural networks have been widely used for natural language processing, time series prediction and many other sequential data tasks. Thus, for these applications there is increasing need for low-power accelerators for LSTM model inference at the edge. In order to reduce power dissipation due to data transfers within inference devices, there has been significant interest in accelerating vector-matrix multiplication (VMM) operations using non-volatile memory (NVM) weight arrays. In NVM array-based hardware, reduced bit-widths also significantly increases the power efficiency. In this paper, we focus on the application of quantization-aware training algorithm to LSTM models, and the benefits these models bring in terms of resilience against both quantization error and analog device noise. We have shown that only 4-bit NVM weights and 4-bit ADC/DACs are needed to produce equivalent LSTM network performance as floating-point baseline. Reasonable levels of ADC quantization noise and weight noise can be naturally tolerated within our NVM-based quantized LSTM network. Benchmark analysis of our proposed LSTM accelerator for inference has shown at least 2.4× better computing efficiency and 40× higher area efficiency than traditional digital approaches (GPU, FPGA, and ASIC). Some other novel approaches based on NVM promise to deliver higher computing efficiency (up to ×4.7) but require larger arrays with potential higher error rates.

## I. Introduction

Deep neural networks [1] have gained great popularity during the past several years and have become one of the most widely used machine learning technique nowadays. Deep neural networks can be broadly classified into two categories: feedforward neural networks and recurrent neural networks, depending on whether there are loops present inside the network topology. Unlike feedforward neural networks such as convolutional neural network (CNN) and multilayer perceptron (MLP) that are being used for static input problems like image recognition, object detection etc., recurrent neural networks such as long short-term memory (LSTM), gated recurrent unit (GRU), and echo state network (ESN) are suitable for non-static input tasks including natural language processing, time-series prediction etc. LSTM [2] is a special kind of recurrent neural networks that was first designed to avoid the exploding or vanishing gradient problems during backpropagation, and have now become the state-of-the-art approach for speech recognition. LSTM, combined with other types of neural networks like CNN, is used by Siri, Google voice, Alexa, etc. but is usually executed on cloud servers and data need to be transmitted between clients and servers through wire/wireless networks which may encounter instability or interruptions. It is desirable to have low-power embedded hardware to run deep neural networks directly on power restricted systems, such as mobile devices and self-driving cars.

Neuromorphic chips are regarded as promising technology to be integrated with mobile devices considering their great advantage in power efficiency and computing speed. They are usually based on CMOS VLSI circuits and attempt to mimic the human brain to perform computations by taking advantage of the massive parallelism when billions of neurons and trillions of synapses process and store information [3]. Some of the existing notable efforts on neuromorphic computing hardware systems include IBM's TrueNorth [4], Stanford's Neurogrid [5], EU's BrainScaleS [6], and more recently Intel's Loihi [7], etc. In addition to using CMOS based analog/digital circuits, non-volatile memory (NVM) devices can be integrated to accelerate neuromorphic computing or machine learning hardware, as they can be used directly as synaptic weights in artificial neural networks [8]. Some of the popular candidate NVM technologies for neuromorphic computing include ReRAM [8], PCRAM [9], MRAM [10] and floating gate transistors [11], which offer smaller footprint and lower power than SRAM or eDRAM that are the mainstream CMOS technologies to hold synaptic weights.

A lot of work has been done previously to investigate NVM based MLP [12]–[14] or CNN [15], [16] hardware accelerators, while only more recently, LSTM acceleration has been reported by using pure analog NVM arrays (ReRAM or PCRAM) [17]–[19]. In [17] the design was realized by using analog NVM, 7-

bit input, and 9-bit ADC, while in [19] 8- or 16-bit number was used to improve the computing efficiency of their LSTM model. In this work, we explore the possibility of using ever lower bit-precision NVM weight array and periphery circuit component, while maintaining comparable performance with the software baseline (32-bit). This is the first work to our knowledge that explore the quantization effects on both NVM weights and ADC/DAC on the periphery for LSTM network, and we show that for different benchmark tasks, 4-bit weight along with 4-bit ADC/DAC is in general able to produce satisfactory result.

The highlights of this paper can be summarized as:

1) Quantization-aware training (potentially performed on CPUs or GPUs) was used for our LSTM network to achieve extreme low bit-precision (< 4 bits) requirements on the NVM array-based hardware for inference purposes, making our network quantization-resilient.

2) Detailed noise analysis on the circuit and device level was performed and it was shown that reasonable amount of ADC quantization noise and weight noise can be naturally tolerated within the quantized LSTM hardware, making our network noise-resilient.

3) Detailed benchmark analysis has shown that our NVM-based LSTM neural network quantized to 4-bit precision can offer 13× higher computing efficiency (throughput / power) and over 8000× higher area efficiency (throughput / area) than state-of-the-art GPU that focuses on edge applications, over 400× higher computing efficiency than FPGA-based LSTM accelerators, 2.4× better computing efficiency and 40× better area efficiency than traditional SRAM-based ASIC approach. Some other NVM-based approaches promise to deliver higher computing efficiency (up to ×4.7) and area efficiency (up to ×3.1) but require larger arrays with potential neural network performance degradation (e.g. higher error rates).

The rest of this paper is organized as follows. Section II discusses the concept of using NVM array to accelerate Vector-Matrix Multiplication (VMM) in machine learning directly using Ohm's law. Section III introduces the basic LSTM operations, how to use NVM arrays to accelerate LSTM operations, and where the quantization should be considered in the hardware implementation. In section IV, quantization bit-widths requirements on the hardware is explored using two major benchmark tasks without considering the analog hardware noise. Section V analyzes the effect of hardware noise on the performance of our quantized LSTM network, and in section VI, detailed benchmark analysis of our NVM-based quantized LSTM network is performed where computing efficiency, area efficiency etc. are compared with mainstream digital approaches (GPU, FPGA, ASIC) and other NVM approaches. Section VII concludes the paper.

## II. Vector-Matrix Multiplication Accelerated by NVM Array

To accelerate machine learning algorithms, the NVM device cells can be constructed into a cross-point like array, as shown in Fig. 1. The cross-point structure is inspired by biology where each pre-synaptic neuron corresponds to each row and each post-synaptic neuron corresponds to each column, therefore each cross junction is one synapse, which is represented by one NVM cell [8]. When used in the read mode, i.e. the conductance values of the NVM weight cells are stationary, the NVM array can accelerate vector-matrix multiplication (VMM) operations directly in physics using Ohm's law [20]. The readout current from each column is the dot product of input voltage values from the rows (when encoded in the amplitude of the voltage pulse) and the stationary conductance values from the NVM cells on that column. Altogether, the readout currents from all the columns represent the VMM of the input voltage vector and the NVM weight array matrix. As VMM is heavily used in most machine learning algorithms, its acceleration efficiency is the most important figure-of-merit in the system.

Such analog VMM realized by using the analog weight array may possibly run into several challenges such as the available NVM cell conductance level is limited to a certain number of bits [21]. Even though ReRAM and PCRAM can have almost continuous incremental conductance change [8], [9], 32-bit precision of weight is hard to achieve, while MRAM and NOR Flash are mostly binary type memory cells [22]. In addition to the inherent limitations posed by the NVM devices, implementing high precision periphery circuits can be very costly in terms of area and power, especially for ADCs due to the number of comparators required for high resolution conversion. Previous studies have shown that ADCs as a classifier between the analog weight array and digital back-end circuits consume most of the power in the system [16]. Therefore, it is important to explore the possibility of using low bit-precision weight memory array and periphery circuit component, while maintaining comparable performance with the software baseline (32-bit).

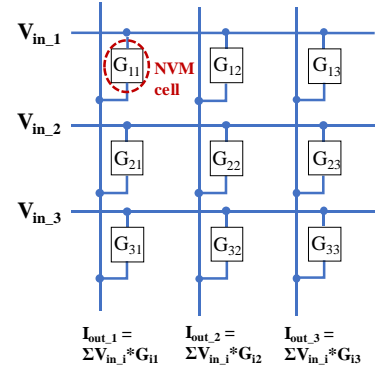

Fig. 1. Schematic of using NVM crosspoint array to accelerate VMM using Ohm's law.

## III. Hardware Accelerated Quantized LSTM

There have been many research efforts from the algorithm point of view on binarizing or quantizing the feedforward neural networks like CNN and MLP [23], [24]. Binarizing LSTM is more challenging than binarizing the CNN or MLP as it is difficult to adopt the back-end techniques like batch normalization in a recurrent neural network [25]. Instead, quantized LSTM has been studied and it is revealed that low quantization bit-widths can be achieved by quantizing the weights and hidden state during forward propagation and using straight-through estimator (STE) to propagate the gradient for weight update [26], [27]. However, these quantized LSTM studies are not based on considerations of the real hardware implementation. For example, hardware implementations based on the NVM weight array need quantization on more than just the weights and hidden state, as will be discussed in the following sections.

*A. Basic LSTM Operations*

The forward propagation operation of the LSTM unit (Fig. 2) contains 4 vector-matrix multiplications, 5 nonlinear activations, 3 element-wise multiplications and 1 element-wise addition [28]. As shown in Equation (1) - (4), the hidden state of the previous time step $h_{t-1}$ is concatenated with the input of the current step $x_t$ to form the total input vector being fed into the weight arrays $W_f$, $W_i$, $W_o$ and $W_c$ to perform the VMM. The VMM results will be passed into 4 nonlinear activation function units respectively to get the values of forget gate $f_t$, input gate $i_t$, output gate $o_t$ and new candidate memory cell $c\_c_t$. The new memory cell $c_t$ is composed of the new information desired to be added by multiplying the new candidate memory $c\_c_t$ with input gate $i_t$, and the old information desired to be not forgotten by multiplying the old memory cell $c_{t-1}$ and forget gate $f_t$, shown in Equation (5). The final hidden state $h_t$ is calculated by multiplying the output gate $o_t$ and the activation of the new memory cell $c_t$, shown in Equation (6). During backpropagation, the values of $W_f$, $W_i$, $W_o$ and $W_c$ are updated according to the training algorithm, usually based on the stochastic gradient descent.

$$f_t = \text{sigmoid}([x_t, h_{t-1}]\ W_f) \quad (1)$$

$$i_t = \text{sigmoid}([x_t, h_{t-1}]\ W_i) \quad (2)$$

$$o_t = \text{sigmoid}([x_t, h_{t-1}]\ W_o) \quad (3)$$

$$c\_c_t = \text{tanh}([x_t, h_{t-1}]\ W_c) \quad (4)$$

$$c_t = f_t \bullet c_{t-1} + i_t \bullet c\_c_t \quad (5)$$

$$h_t = o_t \bullet \text{tanh}(c_t) \quad (6)$$

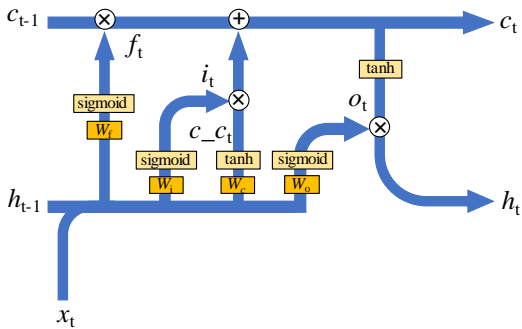

Fig. 2.  LSTM basic operation flowchart

*B. NVM Weight Array Accelerated LSTM Unit*

The 4 vector-matrix multiplications to calculate the forget gate, input gate, output gate and new candidate memory cell can be accelerated by NVM weight arrays, as shown in Fig. 3. The 4 weight arrays representing $W_f$, $W_i$, $W_o$ and $W_c$ can be concatenated into a whole NVM array to calculate the VMM results in parallel. As the input $x_t$ and previous hidden state $h_{t-1}$ processed after the DACs are in the form of analog voltages, NVM weight arrays are resistive cross-point arrays, the VMM results are therefore in the form of analog currents that will pass the ADCs to be converted into digital values. Note that due to the relatively large area of ADCs, it is reasonable to perform time-multiplexing on the ADCs between different columns. Column multiplexers are used to choose which columns to be connected to the ADCs for each time step. After the ADCs, the digital values representing the VMM results will then be fed into different activation function units (either sigmoid or tanh) to get the final values of the forget gate $f_t$, input gate $i_t$, output gate $o_t$ and new candidate memory cell $c\_c_t$ that can later be processed in other hardware components (elementwise multiplication and addition units) to generate the new hidden state $h_t$, which will then be fed into the DAC in the next cycle as part of the total input vector.

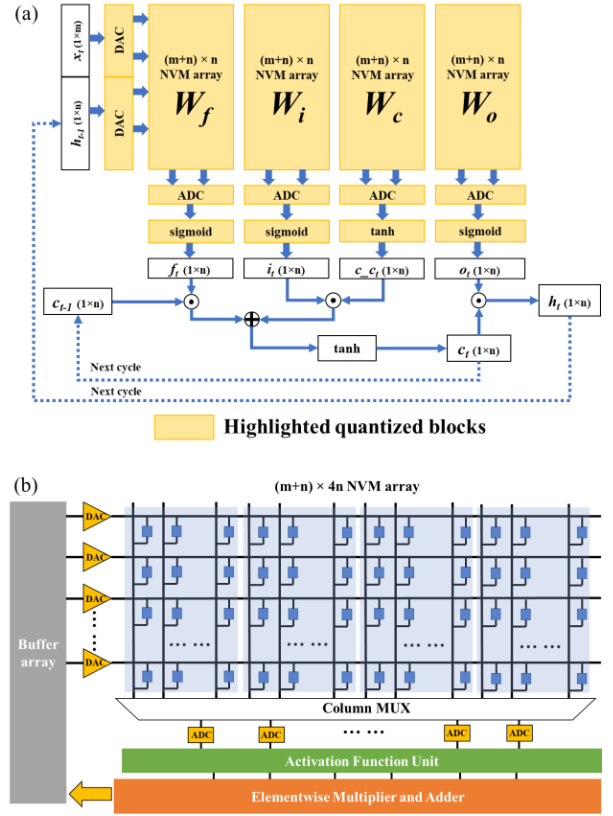

Fig. 3. Architecture of NVM weight array accelerated LSTM unit. (a) Operation flow indicated by block diagram. (b) Hardware design.

As mentioned before, due to the limitations on the available number of stable resistance states on a single NVM cell, it is worthwhile to try to lower the bit-precision of the weights. In addition, the lower bit-precision of the ADC or DAC will lead to lower cost and smaller area/power consumption. Therefore, we target to quantize the highlighted blocks in Fig. 3(a) to a lower bit-width than 32-bit floating-point baseline while we achieve a comparable performance in terms of accuracy. The memory cell values are stored at 32-bit without being quantized to lower bit-width. Note that the activation units (sigmoid or tanh) are quantized naturally with the same bit-widths as the ADCs. Lower bit-width of the non-linear activation units can lead to area/power savings as well. For example, a 4-bit ADC would only require 16 entry look-up tables (LUTs) for sigmoid/tanh activation.

Our training approach is taking quantization into consideration, as the parameters targeted to be quantized, such as weights, inputs from DACs, outputs through ADCs, are already quantized during forward propagation for both training and inference. Straight-through estimator (STE) method is used to propagate the gradients for weight update during back propagation [26], [27]. This quantization-aware training approach is supposed to enhance the performance when the quantization bit-width is extremely low such as 1 or 2 bits. On the contrary, if the quantization is only done offline during inference, such that training is still performed with full precision numbers, it is very difficult to maintain satisfying inference performance with extremely low quantization bit-widths.

## IV. BIT PRECISION REQUIREMENT ON LSTM WEIGHT ARRAY AND CIRCUIT COMPONENTS

To evaluate the performance of our quantized LSTM neural network based on the NVM array, two major tasks are performed: Human Activity Recognition (HAR) and Natural Language Processing (NLP). Different bit-precisions of the weights and ADC/DACs are explored to compare with high precision floating-point baselines. Note that the analog hardware noise is not considered in this section.

### A. Human Activity Recognition (HAR)

The Human Activity Recognition database [29] was built from the recordings of 30 subjects wearing smartphones while performing six daily activities: walking, walking upstairs, walking downstairs, sitting, standing and laying. LSTM is used for this dataset to learn and recognize the type of activity the user is doing. The input vector size and hidden state size for the LSTM unit are both fixed at 32. Therefore, the NVM weight array size used is $64 \times 128$ (m = n = 32 in Fig. 3). Different bit precisions ranging from 1 to 16 bits are used for NVM weights and ADC/DACs and quantization-aware training algorithm using STE is applied.

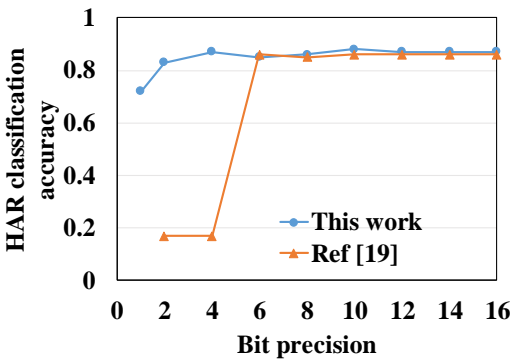

Fig. 4. Human Activity Recognition result. By using the quantization-aware training with straight-through estimator (STE) method, classification accuracy is improved significantly compared to [19] at lower quantization bit-widths such as 1-4 bits.

To compare this result with a similar work [19] that uses retraining-based iterative parameter quantization, where training is still carried out with full precision numbers and then followed by quantization and retraining for each iteration, we observe a significant improvement on the classification accuracy at lower quantization bit widths such as 1 – 4 bits (Fig. 4). It is important to point out that with 4-bit NVM weight precision and 4-bit ADC/DAC, the classification accuracy does not degrade compared to the 16-bit baseline.

### B. Natural Language Processing (NLP)

Two subtasks are chosen for NLP. The first subtask - the Penn Treebank dataset [30] contains 10K unique words from Wall Street Journal material annotated in Treebank style. With the Penn Treebank corpus, the task is to predict the next word in the sentence based on previous words, and the performance is measured in perplexity per word (PPW). The perplexity is roughly the inverse of the probability of correct prediction. The input vector size and hidden state size are both fixed at 300. Therefore, the NVM weight array size is $600 \times 1200$ (m = n = 300 in Fig. 3). As shown in Fig. 5, as training progresses, the validation perplexity continues decreasing for the 32-bit FP baseline, 2-bit weight 2-bit ADC/DAC, and 4-bit weight 4-bit ADC/DAC cases. The 1-bit weight 2-bit ADC/DAC example case does not show a successful training as the validation perplexity does not converge, while the 4-bit weight 4-bit ADC/DAC case produces competitive training result with the FP baseline without noticeable degradation.

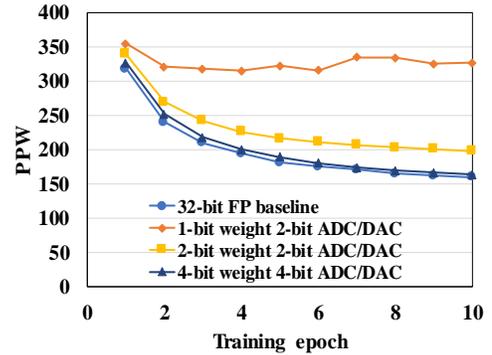

Fig. 5. Penn Treebank dataset result. Validation perplexity does not converge for the 1-bit weight 2-bit ADC/DAC case while the other bitwidth configurations produce successful training. 4-bit weight 4-bit ADC/DAC can generate close-to-equivalent training result with the FP.

To fully explore the bit-width requirement on the weights and ADC/DAC, all combinations of bit-precision ranging from 1 to 4 bits are tested as shown in Fig. 6, 4-bit weight along with at least 2 bits of ADC/DAC are required to achieve a comparable result with the floating-point baseline (less than 5% of perplexity increase). It clearly turns out that the high bit-precision of the weight plays a more important role than the high bit-precision of ADC/DAC for the general performance of the LSTM network, such as with 1-bit weight and 2-bit ADC/DAC the final PPW is 327, which is higher than 282, the PPW achieved with 2-bit weight and 1-bit ADC/DAC. Similar phenomenon was observed by comparing performances between the 2-bit weight 4-bit ADC/DAC case (PPW=182) and the 4-bit weight 2-bit ADC/DAC case (PPW=172). Therefore, we can conclude that improving the resolution of the conductance levels of NVM cell weights is more important than increasing the precision of ADC/DAC.

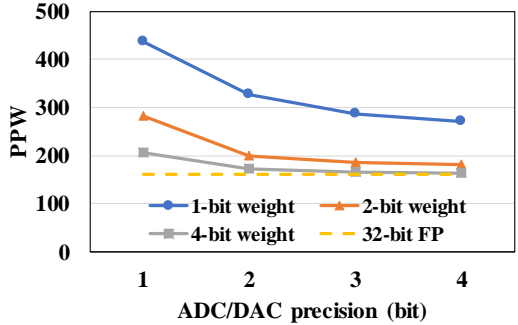

Fig. 6. Penn Treebank dataset result with full exploration on the bit-widths of weight and ADC/DAC. The PPW is measured as the validation perplexity after 10 epochs of training.

The second subtask is the national name prediction where the next character in a name is predicted instead of the next word. The perplexity metric here is for per character. The input vector size is 100 while the hidden state size is 256. So, the NVM weight array size used for this task is 356 × 1024 (m = 100, n = 256 in Fig. 3). After 8000 training iterations, the training perplexity and accuracy are measured. As shown in Table I, using 2-bit weight and 2-bit ADC/DAC is sufficient to produce result within 5% degradation compared to the floating point baseline case. Comparing with the result from the Penn Treebank, a lower bit-precision requirement on the weight and ADC/DAC is needed for the simpler character prediction task. To conclude from both NLP tasks, a 4-bit weight along with 4-bit ADC/DAC can ensure almost-zero degradation for LSTM network performance. Such bit-width requirements for training also ensure the inference performance as our training approach is quantization-aware. Quantization-aware model-based training ensures that the forward pass matches precision for both training and inference.

TABLE I. TRAINING PERPLEXITY AND ACCURACY AT DIFFERENT BIT-WIDTHS CONFIGURATIONS FOR CHARACTER PREDICTION

| Character prediction result | | |
|---|---|---|
| LSTM configuration | Training accuracy (%) | Training perplexity (per character) |
| 32-bit floating point baseline | 85.09 | 1.52 |
| 4-bit weight + 4-bit ADC/DAC | 85 | 1.55 |
| 2-bit weight + 2-bit ADC/DAC | 83.6 | 1.58 |
| 1-bit weight + 1-bit ADC/DAC | 72.82 | 2.27 |

## V. EFFECT OF DEVICE AND CIRCUIT NOISE

After we show the possibility of using lower bit-precision weights and peripheries for LSTM, non-idealities in the quantization precisions of NVM weight cells and ADC/DACs should be considered for their potential impact on network performance. Especially during forward propagation, the read noise can distort the VMM result and thus lead to inaccurate weight updates. The two main sources of noise are the analog digital conversion and the device itself: the resistance of the memory element is intrinsically fluctuating. Specifically, the types of noises being considered include:

1) Johnson-Nyquist (thermal) noise, which is caused by the random thermal motion of charge carriers. The effect is intrinsic to any resistor and puts a lower limit on how fast any memory element can be read.

2) Shot noise, which is the statistical fluctuation of the electric current due to its quantized nature, and manifests predominantly when the current traverses a barrier, like in many resistive memory cells.

3) Random telegraph noise (RTN), or burst noise, which is commonly caused by the random motion of charge carriers between trapping sites. The output signal typically appears to switch back and forth between different levels.

4) 1/f (flicker) noise, which has a 1/f power spectral density and can be the result of impurities, generation and recombination, and randomly distributed traps. Superposition of many RTN processes on different scales will have a 1/f spectrum.

5) Quantization noise, which is the error introduced by quantization in the analog-to-digital converter (ADC). It is the rounding error between the analog input voltage of the ADC and the output digitized value and is discussed as follows.

### A. Effect of ADC Noise

The ADC noise consists of thermal as well as quantization noise. For relevant bandwidths, the quantization noise, $V_{\text{QNoise}} = \Delta/\sqrt{12}$ [31], with $\Delta$ being the quantization step size of ADC, is larger than thermal and shot noise. For example, for a signal range of ±1 V and 2-bit resolution we can expect a quantization noise of $V_{\text{QNoise}} = 1/(2^1 \cdot \sqrt{12}) = 0.144$V, which is dominant compared to the Johnson-Nyquist noise $V_{\text{JNoise}} = \sqrt{k_B T \cdot R \cdot \text{BW}}$ [32], which is around 4 nV / $\sqrt{\text{Hz}}$ for a 1 k resistor. Therefore, the resistive thermal noise of the ADC can be neglected.

Another way to look at the noise is to calculate the effective number of bits from the noise power in relation to the signal power. Effective number of bits, or ENOB = (SNR – 1.76dB) / 6.02dB [31], with signal to noise ratio (SNR) defined as $P_{\text{signal}}/P_{\text{noise}}$. This is essentially the same as the equation for the root mean square value of $V_{\text{QNoise}} = \Delta/\sqrt{12}$.

To simply model the ADC quantization noise, an additive noise term is added to the values at the forget gate, input gate, output gate and new candidate memory cell during ADC quantization and before the activation function units. The noise follows a Gaussian distribution with a standard deviation $\sigma = V_{\text{QNoise}} = (V_{\max} - V_{\min})/(2^N \cdot \sqrt{12})$, $N$ is the ADC bit resolution. For example, at the forget gate:

$$f_t = sigmoid\ ([x_t, h_{t-1}]\ W_f + Z_{\text{ADC}}) \qquad (7)$$

$$Z_{\text{ADC}} \sim N(0, \sigma^2),\ \sigma = V_{\text{QNoise}} \qquad (8)$$

Where $Z_{ADC}$ is the ADC quantization noise vector with the same dimension as $[x_t, h_{t-1}] W_f$. It follows a Gaussian distribution with zero mean and a standard deviation $\sigma$ that equals the ADC quantization noise root mean square value, which is $V_{QNoise}$ by definition. As shown in Fig. 7, the influence of ADC quantization noise on the LSTM performance is negligible. The experiment is run on the Penn Treebank corpus measuring the validation perplexity after 10 epochs of training. Note that the noise is also present during the quantization-aware training stage as our LSTM network can be considered for online training as well.

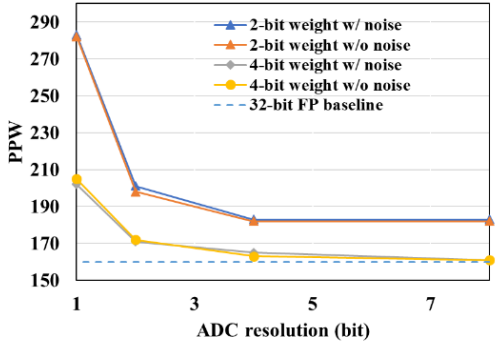

Fig. 7. Effect of ADC quantization noise on the Penn Treebank experiment. Weight resolution are fixed at either 2 bits or 4 bits. Validation perplexity is measued.

### B. Effect of Weight Noise

As explained in the previous section, the resistive thermal (Johnson-Nyquist) noise will also be negligible for the memory cells, as well as the shot noise, which is given by $I_{SNoise} = \sqrt{2e \cdot I \cdot BW}$ [32]. Due to the atomistic nature of transport in resistive memory cells, $1/f$ noise and random telegraph noise (RTN) will be dominant [33]. From literature, we know that the level of RTN noise and $1/f$ noise is dependent on the resistance levels, such as $\sigma_R/R$ is around 0.1 at high resistance (low conductance) states and around 0.01 at low resistance (high conductance) states.

To model the weight noise, an additive noise term is added to the values of the weight arrays. The noise follows a Gaussian distribution with a standard deviation proportional to the total weight range. For example, at the forget gate:

$$f_t = sigmoid\ ([x_t, h_{t-1}]\ (W_f + Z_w)) \qquad (9)$$

$$Z_w \sim N\ (0,\ \sigma^2),\ \sigma = \beta\ (w_{max} - w_{min}) \qquad (10)$$

Where $Z_w$ is the weight noise matrix with the same dimension as $W_f$. It follows a Gaussian distribution with zero mean and a standard deviation $\sigma$ ranging from 0 to 20% of the total weight range $w_{max} - w_{min}$. We define the percentage of the weight range $\beta$ as the weight noise ratio. Only exploring $\beta$ from 0 to 20% is realistic with previously mentioned NVM device performance. Fig. 8 shows that even 20% weight noise have no harmful effect on the LSTM network performance with the same Penn Treebank experiment setup. From analyzing both the ADC quantization noise and weight noise, we can clearly say that our LSTM network is robust against reasonable levels of hardware noise.

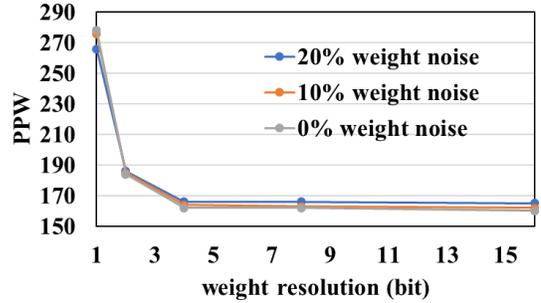

Fig. 8. Effect of weight noise on the Penn Treebank experiment. ADC/DAC resolution is fixed at 4 bits. Validation perplexity is measured after 10 epochs of training.

## VI. BENCHMARK ANALYSIS

Here we compare the inference performance of our NVM-based quantized LSTM network at 4-bit weight and 4-bit ADC/DAC configuration with other existing hardware platforms including GPU (Nvidia Jetson AGX Xavier [34]) that focuses on edge applications, FPGA-based LSTM accelerator [35] that uses DRAM for weight storage, ASIC-based neuromorphic engine [36] that uses SRAM for weight storage and other NVM-based LSTM accelerators [17], [19]. For GPU- and FPGA-based systems, the high throughput usually comes with high power and area consumption, so the overall computing and area efficiency are generally lower than ASIC- and NVM-based systems (Table II, Table III, and Fig. 9).

### A. Throughput Analysis

The latency for NVM based vector-matrix multiplication is around 80 - 100 ns [17], [19]. Therefore, for an NVM array size of $356 \times 1024$ used in our paper for the character prediction task, the estimated throughput for vector-matrix multiplication is 3645 GOP/s (assuming 100 ns latency with 64 ADCs at sampling rate of 160 MS/s), which is comparable with GPU. In addition, we need to consider the latency for non-linear activation function, which is assumed to be 5 ns, and element-wise operations, which is assumed to be 1 ns. Communication overhead is not considered here. So, the overall throughput is calculated to be 3439 GOP/s. IBM estimates that their resistive processing unit (RPU) devices [17] accelerate the throughput to be 84 TOP/s and assume 3 resistive cross-point arrays with the size of $4096 \times 4096$ are used at the same time. In general, larger array size enables higher throughput, but the line resistance and parasitic capacitance of larger arrays can become non-negligible and will potentially harm the network accuracy. Note that in the RRAM-based Processing-In-Memory (PIM) architecture paper [19], each subarray size is only $128 \times 128$, thus the achieved throughput is only 108.4 GOP/s.

### B. Power Estimation

ADCs can consume a relatively large amount of power in the overall architecture. For a $356 \times 1024$ NVM array it is unrealistic to use 1024 ADCs with full parallel readout for all the columns. Therefore, we locate 64 ADCs to perform time-multiplexing so that every 16 columns are sharing one ADC (for the 356 × 1024 array). Suppose we need 100 ns latency for each column, so the sampling rate of ADC needs to be at least 160 MS/s. Energy consumption per sample in 4-bit ADC is assumed as 1 pJ [37]. For ADCs at lower resolutions, the energy performance seems to be independent of ENOB. For higher resolutions such as ENOB ≥ 9, the energy per sample quadruples for every additional bit of effective resolution in ADC. The state-of-the-art energy per sample vs. ENOB almost exactly follows the relationship $E = 2^{2(ENOB-9)}$ pJ for ENOB ≥ 9. Therefore, in our design with 4-bit ADC with 160 MS/s sampling rate, each ADC consumes 0.16 mW, and with 64 4-bit ADCs the estimated power is 0.01 W. On the other hand, if 12-bit ADCs are used instead, the energy consumption per sample will skyrocket to 64 pJ and the overall power consumption from 12-bit ADCs will be 0.64 W.

TABLE II. COMPARISON WITH MAINSTREAM DIGITAL APPROACHES

|  | GPU Nvidia Jetson [34] | ESE [35] | Tianjic [36] | This work |
|---|---|---|---|---|
| Technology | GPU | FPGA | ASIC | NVM |
| Precision (bit) | 16 | 12 | 8 | 4 |
| Throughput (GOP/s) | 3,478 | 282 | 1,214 | 3,439 |
| Power (W) | 15 | 41 | 0.95 | 1.136 |
| Computing efficiency (GOP/s/W) | 231 | 6.88 | 1,278 | 3,027 |
| Area (mm²) | 8700 | N/A | 14.44 | 1.031 |
| Area efficiency (GOP/s/mm²) | 0.399 | N/A | 84 | 3,333 |

TABLE III. COMPARISON WITH OTHER NVM-BASED APPROACHES

|  | IBM's RPU [17] | RRAM PIM [19] | This work |
|---|---|---|---|
| Array size | 4096 × 4096 | 128 × 128 | 356 × 1024 |
| Precision (bit) | N/A | 16 | 4 |
| Throughput (GOP/s) | 84,000 | 108.4 | 3,439 |
| Power (W) | 6 | 0.932 | 1.136 |
| Computing efficiency (GOP/s/W) | 14,166 | 116.3 | 3,027 |
| Area (mm²) | 8.04 | 0.39 | 1.031 |
| Area efficiency (GOP/s/mm²) | 10,477 | 277 | 3,333 |

For estimating read power consumption in NVM array while performing inference operation, the assumed read voltage for our NVM cell is 1 V, and the assumed average resistance of the NVM cells is 1 MΩ. Therefore, the power from the 356 × 1024 NVM array is 0.364 W.

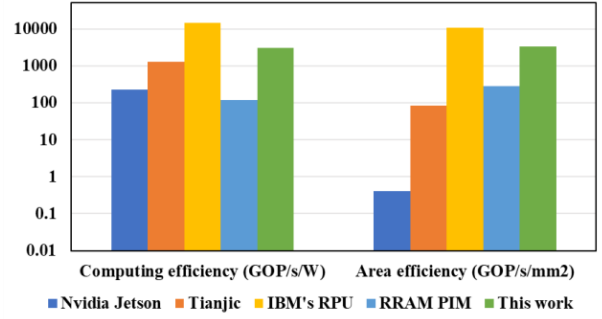

Fig. 9. Comparison of our NVM-based quantized LSTM neural network with other hardware platforms.

In addition to the power consumption by the ADCs and NVM array, we need to consider how much power the other peripherals including the DACs, multiplexers, data buffers, non-linear activation function units, and element-wise operations consume. We utilized relevant numbers from [19] and came to a total number of 1.136 W for power consumption of our design.

The computing efficiency of our NVM-based quantized LSTM (throughput divided by power) – around 3 TOP/s/W is higher than traditional digital approaches (Table II). FPGA platforms show relatively lower computing efficiencies because of their relatively higher power consumption (41 W) and lower throughput (282 GOP/s), while Nvidia Jetson GPU and ASIC-based approaches are in the middle range of performance. Compared with other NVM-based approaches, our computing efficiency is in the middle (Table III). IBM's RPU shows exceedingly high computing efficiency due to the enormous size of the array being used. Note that if the neural network precision in [19] is decreased to 4-bit, their computing efficiency is estimated to 2.7 TOP/s/W, which is very close to our result. Therefore, we see the obvious advantage of utilizing low bit-width to enhance computing efficiency.

*C. Area Estimation*

Similar to the power estimation, the large footprint of ADC in nature can be dominant in overall system area. The area of ADC increases by 2-2.2× for every additional bit of ENOB when ENOB ≥ 6 [38]. A 4-bit ADC consumes around 0.01 mm² [38] so that the estimated area of 64 4-bit ADCs on a 356 × 1024 array is 0.64 mm². In comparison, if 12-bit ADCs are used instead, the area consumption per ADC will increase to 1 mm². Therefore, the overall area consumption from 12-bit ADCs will be 64 mm², which is intolerable.

We assume that the NVM array itself has 400 nm pitch with 200 nm NVM device width and 200 nm spacing between devices. A 356 × 1024 array will therefore consume 0.058 mm², which is insignificant compared to the ADC area consumption. In addition to ADCs and NVM array, area consumption by the DACs, multiplexers, data buffers, non-linear activation function units, and element-wise operations cannot be neglected. Similarly, we utilized relevant numbers from [19] and came to a total of 1.031 mm² for area consumption of our design.

The area efficiency of our NVM-based quantized LSTM (throughput divided by area) – around 3 TOP/s/mm² is also among the highest in the benchmark analysis as shown in Table

II, Table III, and Fig. 9. By lowering the resolution of ADCs, significant amount of power and area can both be saved in our design. As for the NVM weights, the bit-width is not critical as much as ADC's for our current design in terms of power and area performance as the NVM memory cells are inherently multi-level cells.

## VII. CONCLUSION

Quantizing the LSTM neural network becomes very important when it comes to the embedded hardware design and acceleration of state-of-the-art machine learning to lower the memory size and computation complexity. Specifically, NVM cross-point arrays can accelerate the VMM operations that are heavily used in most machine learning algorithms for artificial neural networks, including but not limited to LSTM, CNN and MLP. Previous literature has reported on using pure analog NVM arrays for LSTM without considering any quantization bit-width requirements on the weight memory cells or circuit components [17], [18]. To our best knowledge, this is the first work that explore the quantization effects on both NVM weights and ADC/DAC on the periphery for LSTM.

By utilizing a quantization-aware training approach, our NVM-based LSTM network is both quantization- and noise-resilient. We have found that 4-bit NVM weight cell along with 4-bit ADC/DAC in the LSTM unit can deliver comparable performance as the floating-point baseline in human activity recognition and natural language processing tasks. For a simple dataset for character level prediction, even 2-bit NVM weight cell along with 2-bit ADC/DAC does not show noticeable degradation in performance. In addition, ADC quantization noise and NVM weight noise can both be well tolerated for our quantized LSTM network as well, making it robust against realistic hardware noise.

Detailed benchmark analysis has shown that our NVM-based quantized LSTM neural network at 4-bit precision can offer at least 13× higher computing efficiency and 8000× times higher area efficiency than Nvidia Jetson GPU, over 400× times higher computing efficiency than the FPGA-based LSTM accelerator – ESE, 2.4× better computing efficiency and 40× better area efficiency than the neuromorphic chip – Tianjic. Some other novel approaches based on NVM, such as IBM's RPU, promise to deliver higher computing efficiency (up to ×4.7) but require larger arrays (4096 × 4096) with potential higher error rates caused by the non-negligible parasitic capacitance and line resistance. Even with competitive computing efficiency, traditional ASIC-based method still has relatively low area efficiency due to the larger area of SRAM devices, while NVM-based approaches in general already start to show area efficiency advantage.

## ACKNOWLEDGMENT

The authors would like to thank Adel Dokhanchi, Tung Hoang and Haoyu Wu for useful discussions.

*arXiv:1805.11801*, 2018.